\documentclass{article}
\pdfoutput=1

\usepackage[numbers]{natbib}
\usepackage[final]{neurips_2019}

\usepackage[utf8]{inputenc} 
\usepackage[T1]{fontenc}    
\usepackage{hyperref}       
\usepackage{url}            
\usepackage{booktabs}       
\usepackage{amsfonts}       
\usepackage{nicefrac}       
\usepackage{microtype} 
\usepackage{graphicx}
\usepackage{subfigure}
\usepackage{multirow}
\usepackage[table,xcdraw]{xcolor}

\title{Establishing an Evaluation Metric to \\ Quantify Climate Change Image Realism}

\author{%
 Sharon Zhou\thanks{equal contribution}\\
  Stanford University\\
  \texttt{sharonz@cs.stanford.edu} \\
  \And
  Alexandra Luccioni\footnotemark[1]\\
  Mila, Université de Montréal\\
  \texttt{luccionis@mila.quebec} \\
  \And
  Gautier Cosne\\
  Mila, Université de Montréal\\
  \texttt{cosnegau@mila.quebec} \\
  \And
  Michael S. Bernstein \\
 Stanford University\\
  \texttt{msb@cs.stanford.edu} \\
  \And
  Yoshua Bengio \\
 Mila, Université de Montréal\\
  \texttt{yoshua.bengio@mila.quebec}
}

\begin{document}

\maketitle

\begin{abstract}
With success on controlled tasks, generative models are being increasingly applied to humanitarian applications~\cite{medicalgan, deepempathy}. In this paper, we focus on the evaluation of a conditional generative model that illustrates the consequences of climate change-induced flooding to encourage public interest and awareness on the issue.
Because metrics for comparing the realism of different modes in a conditional generative model do not exist, we propose several automated and human-based methods for evaluation. To do this, we adapt several existing metrics, and assess the automated metrics against gold standard human evaluation. We find that using Fr\'echet Inception Distance (FID) with embeddings from an intermediary Inception-V3 layer that precedes the auxiliary classifier produces results most correlated with human realism. While insufficient alone to establish a human-correlated automatic evaluation metric, we believe this work begins to bridge the gap between human and automated generative evaluation procedures.
\end{abstract}

\section{Introduction}
\label{introduction}

\begin{figure}[h!]
\vskip 0.2in
\begin{centering}
 \begin{subfigure}{}
   \textbf{High Realism} \\
    \includegraphics[width=0.3\columnwidth]{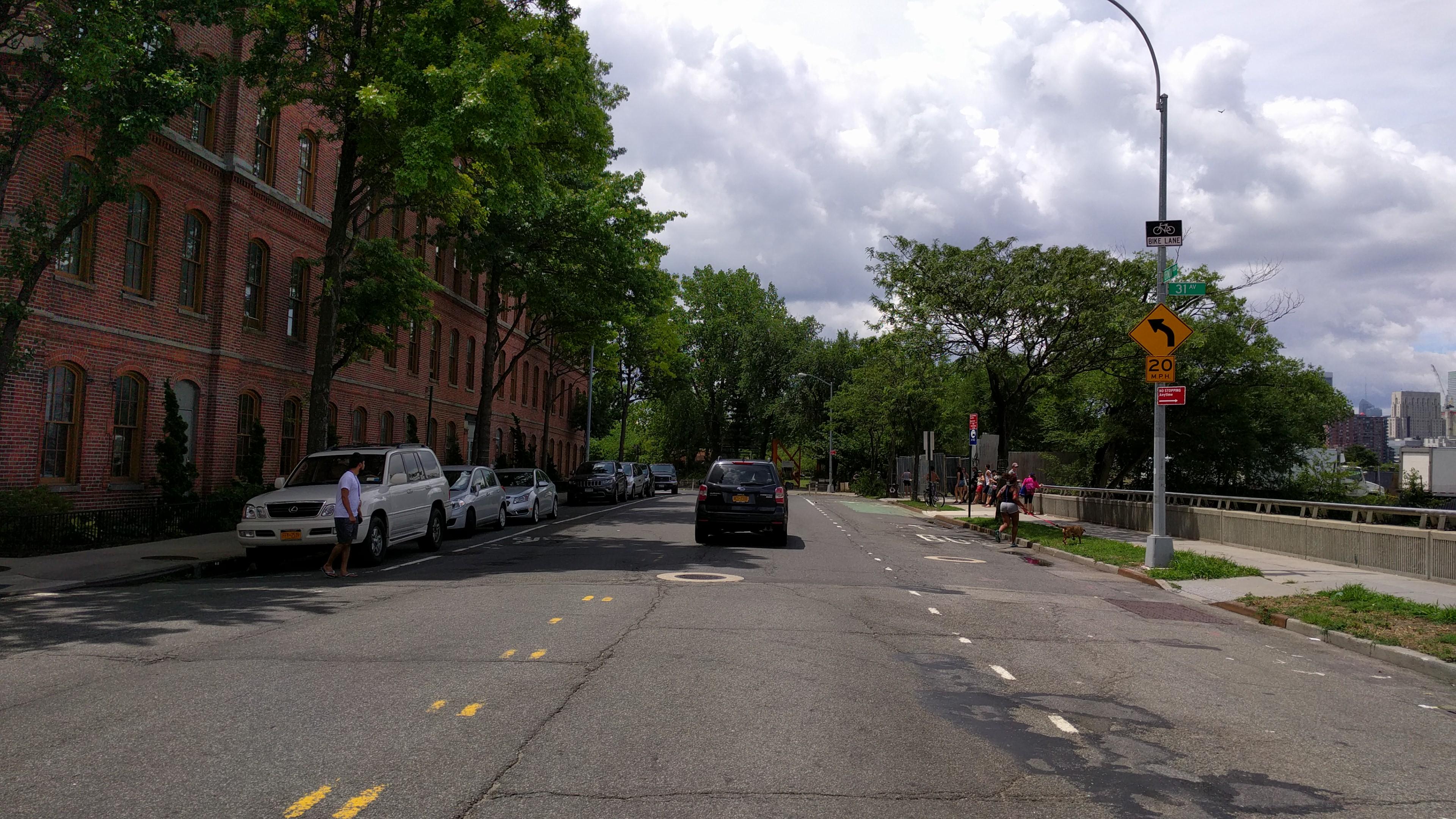}
  \end{subfigure}
  \begin{subfigure}{}
   \includegraphics[width=0.3\columnwidth]{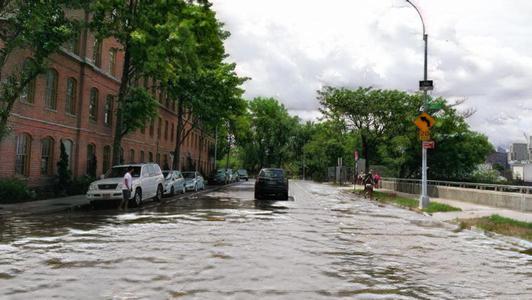}
  \end{subfigure}
  \\
    \textbf{Average Realism}\\
   \begin{subfigure}{}
    \includegraphics[width=0.3\columnwidth]{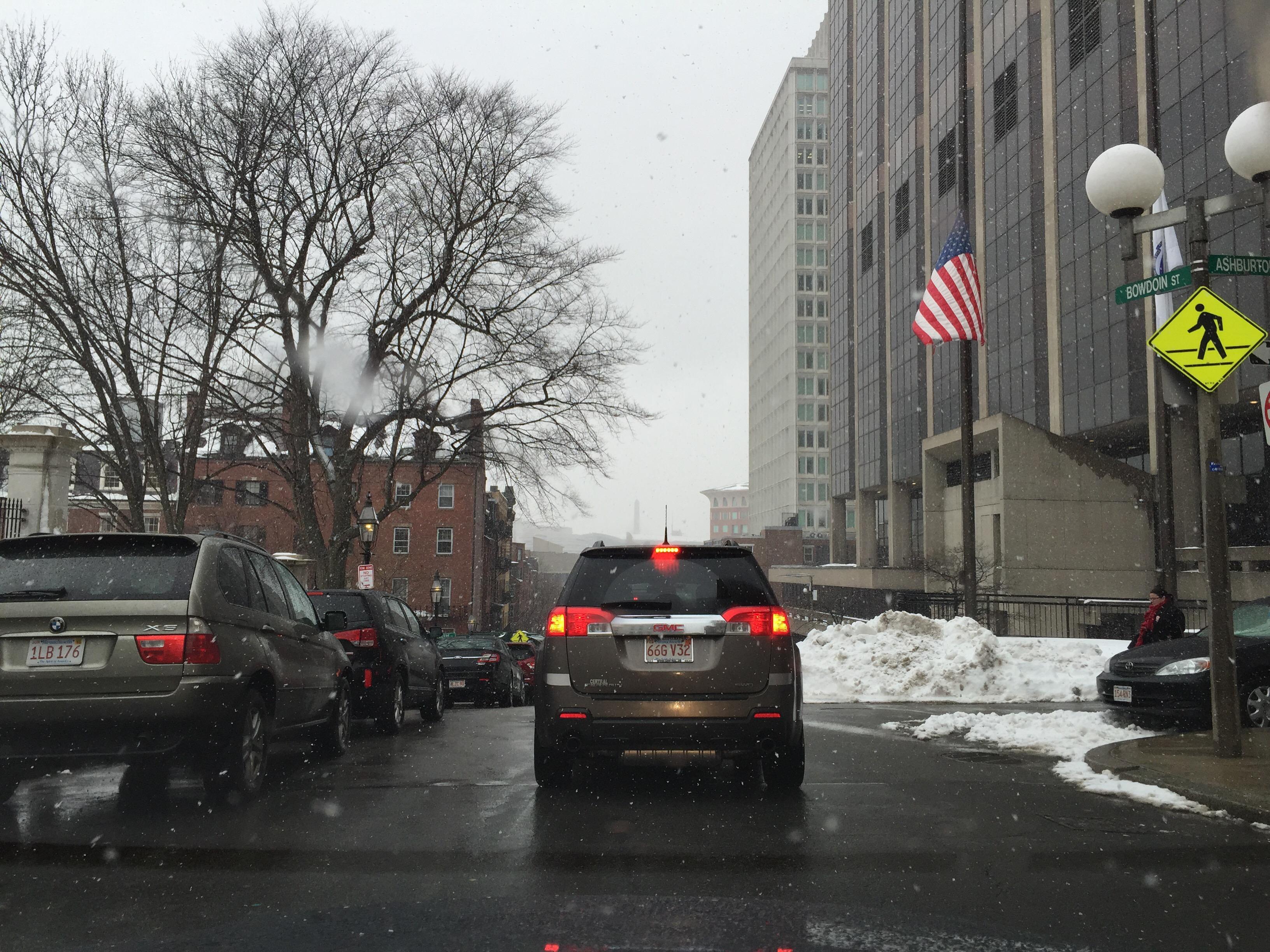}
  \end{subfigure}
  \begin{subfigure}{}
    \includegraphics[width=0.3\columnwidth]{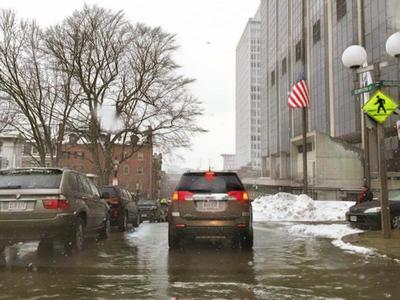}
  \end{subfigure}\\
  \textbf{Low Realism} \\
  \begin{subfigure}{}
    \includegraphics[width=0.3\columnwidth]{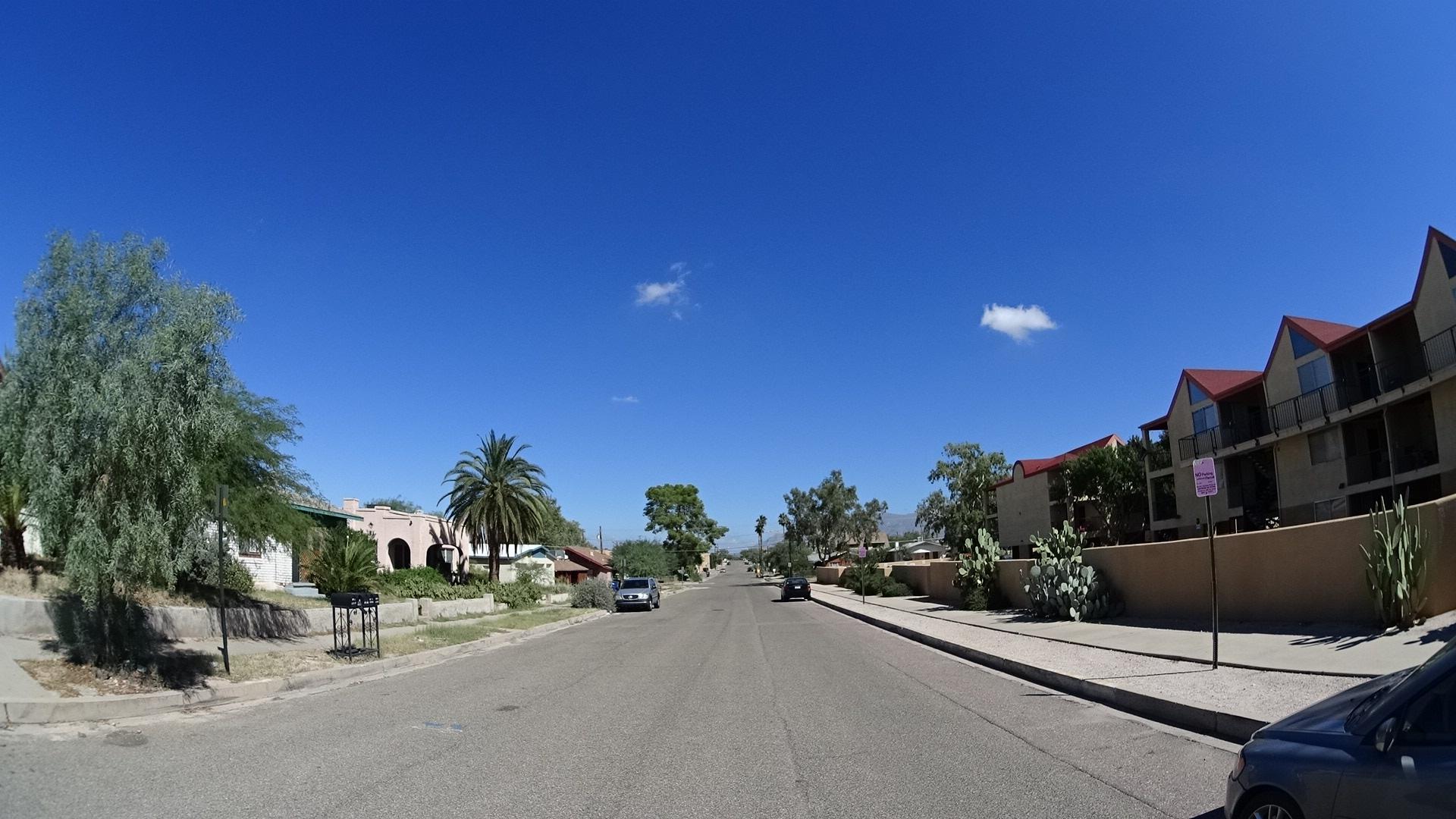}
  \end{subfigure}
  \begin{subfigure}{}
    \includegraphics[width=0.3\columnwidth]{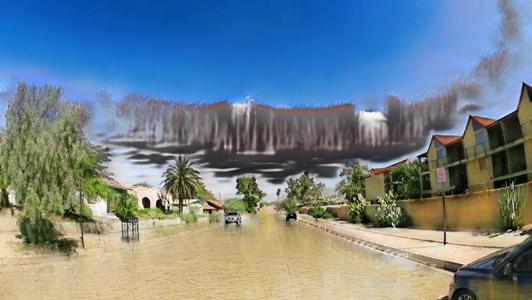}
  \end{subfigure}
  \caption{Generated images of flooded scenes from our model on the right (input on the left), spanning a range of scenes including an urban street, a city view, and a suburban scene. These images also span a range of performance on human perceptual realism, from highly realistic on top to highly non-realistic on the bottom.}
  \label{flooding}
\end{centering}
\end{figure}

Historically, climate change has been an issue around which it is hard to mobilize collective action, notably because public awareness and concern around it does not match the magnitude of its threat to our species and our environment~\cite{pidgeon, weber}. One reason for this mismatch is that it is difficult for people to mentally simulate the complex and probabilistic effects of climate change, which are often perceived to be distant in terms of time and space~\cite{cc-iconic}. Climate communication literature has asserted that effective communications arises from messages that are emotionally charged and personally relevant over traditional forms of expert communication such as scientific reports~\cite{cc-comms}, and that images in particular are key in increasing the awareness and concern regarding the issue of climate change~\cite{cc-images}. With this in mind, our project leverages the MUNIT architecture~\cite{munit} to perform cross-domain multimodal mapping between a street-level image without any flooding to multiple versions of this image under diverse flood styles, to visually represent the impact of climate change-induced flooding on a personal level (for results of our model, see Figure~\ref{flooding}).

Generative models suffer from a lack of strong evaluation methods for mode comparison. Undeniably, much of the utility of generative models arises from their ability to produce diverse, realistic outputs, in addition to controlling generation---such as over specific modes, class labels~\cite{cgans}, or visual attributes~\cite{cgans-visual}---using conditional constraints. Conditional GANs have two inputs: the conditioning input (in our case the image of a house which is not flooded) and the random noise $Z$ which selects a \textit{style}, or mode of the conditional distribution. Existing methods for evaluating the quality and diversity of the generated outputs have strong limitations, and are particularly scarce for conditional models. Widely used metrics include using heuristic approximations~\cite{stylegen,inception,fid} that do not necessarily correlate with human judgment~\cite{hype}, rendering quantitative measurement of progress difficult. We encountered this problem during the course of the development of our model and in this paper propose methods for quantifying the realism of modes learned by a generative model, starting with our own human evaluation. Following this, we compare human and automated approaches for evaluating the output of multimodal generative models, illustrated in the context of our image generation task.

\section{Related Work}
\label{related}
To date, there have been two main approaches proposed for generative model evaluation: automated metrics such as Kernel Inception Distance (KID) ~\cite{Binkowski2018DemystifyingMG}, Inception Score (IS)~\cite{inception} and Fr\'echet Inception Distance (FID)~\cite{fid}, which all aim to evaluate both the visual quality and sample diversity of generated samples at the distribution level, and, more recently, human-centered metrics such as HYPE (Human eYe Perceptual Evaluation)~\cite{hype}, which use human evaluators to assess image realism. Both approaches have their advantages and drawbacks: while automated metrics are cheap and easy to compute, they need large sets of both generated and real samples in order to produce reliable scores, which even then are not comparable between different tasks. Human metrics, on the other hand, may be more representative of human perception, but are more costly to compute and can vary depending on task design~\cite{crowdsourced, comparing}.

Recent work has proposed ways of extending existing automated metrics, for instance by using a modified version of FID for conditional models~\cite{fjd} and sampling heuristics such as the truncation trick~\cite{truncation}. However, these modifications do not evaluate the visual fidelity between different modes, only within them in the case of Fr\'echet Joint Distance~\cite{fjd}, which limits their application in multimodal settings such as ours. Methods for detecting artifacts~\cite{artifacts} and artificial fingerprints~\cite{marra2019gans} in generated samples also touch on perceptual fidelity, but are either, in the case of artifacts, a subset of image realism or, in the case of artificial fingerprints, encompass non-perceptual qualities that are imperceptible to a human viewer. Therefore, within the scope of our research, we found no satisfactory automated metric that would allow us to evaluate the realism of the images that we generated, and we endeavoured to find new ways of doing so, which we describe below.

\section{Evaluating Image Realism}
The research questions that we aim to answer are as follows: (1) What is the most effective way to evaluate the realism of different styles generated by our model? (2) Can we propose an automated method that is correlated with human perceptual realism for automatically selecting the best mode on the flood generation task? We frame this task at the style level: for each given style vector, we aggregate across multiple samples conditioned on the same mode. This style-level aggregation avoids evaluating on individual samples, which would produce noisier comparisons. We accomplish this by adapting the HYPE metric for style-level assessment using crowdsourced human evaluation, and call our new metric \textit{HYPE-Style} (see Section~\ref{experiment}). We compare HYPE-Style against various automated metrics. These metrics adapt FID and KID to the style comparison task. For each metric, we also experiment with different Inception layers.

We analyze Pearson's correlation coefficient $r$ between each proposed automated style ranking method and HYPE-Style to identify the method most correlated with human perceptual realism. The measure $r$ has support [-1,1], where values of 1 and -1 indicate strong positive and negative correlation, respectively, while values around zero indicate low correlation. An $r$ of 1 is the maximum performance achievable on this metric. We also compute the 95\% bootstrapped confidence intervals (CIs) on $r$ using 25 replicates in order to determine the separability of the scores. For each replicate $i$, we compute HYPE-Style and an automated score using images sampled with replacement, from which we calculate $r_i$. We report the median $r$ values, with 95\% bootstrap CIs. 

\subsection{HYPE-Style: Human Evaluation} \label{experiment}
In order to establish a human gold standard, we evaluated 500 image-style combinations drawn from our model, based on 25 input images of diverse locations and building types (houses, farms, streets, cities), each with 20 styles generated by our model. To establish the human baseline, we presented 50 images to each of our human evaluators: 25 real flooded images and 25 generated images. Following prior work, evaluators were calibrated and filtered by this tutorial of half real and half generated images, and were given unlimited time to label an image real or fake~\cite{hype}. 
For each image, we compute the average error rate, which corresponds to the proportion of human evaluators who judged the image as real. Higher values indicate more realistic images. 

\begin{table*}[t]
\centering
\scalebox{0.92}{
\begin{tabular}{|cc|c|c|c|c|}
\hline
\multirow{3}{*}{} &     & \textbf{pool~1}                 & \textbf{pool~2}                 & \textbf{pre-aux}              & \textbf{pool~3}                \\
\hline
 & \textbf{FID} & 0.103 (0.53, 0.153)   & 0.146 (0.099, 0.193)  & \textbf{0.433 (0.390, 0.476)} & 0.407 (0.366, 0.448) \\
 & \textbf{KID} & 0.010 (-0.041, 0.061) & 0.034 (-0.015, 0.083) & 0.432 (0.389, 0.475) & 0.367 (0.322, 0.412) \\
\hline
\end{tabular}}
\centering
\caption{\textbf{Pearson's $r$ Correlation Coefficient}. Results of Pearson's $r$ and bootstrap 95\% confidence intervals between human HYPE-Style scores and all automated methods across different layers of an ImageNet-pretrained Inception-V3 model, including the three pooling layers (\textit{pool~1}, \textit{pool~2}, \textit{pool~3}) and the layer preceding the auxiliary classifier (\textit{pre-aux}). Higher values indicate greater correlation.}
\label{result-table}
\end{table*}

We make several modifications to prior work in order to enable intra-style comparisons in conditional generation. Instead of randomly sampling across all generated images, we constrain the procedure in two ways:~(1) we require that each style and image combination is evaluated multiple times, so we have comparisons between styles yet still within a given image, and~(2) we ensure that evaluators do not see multiple styles generated from a given input image, as this visual redundancy would reveal that they were generated. These two adaptations increase the number of evaluators needed for this task, as evaluators are restrained to a limited set of images sans input redundancy, while still needed to evaluate across different styles for given input images. 

We also diverge from HYPE when calculating scores, aggregating images by style into groups and computing the micro-average of all human evaluator labels within each group. Specifically, for each style $s$ and image $x$, we have multiple human labels $l_x^s$ marked either ``real" (1) and ``generated" (0) based on human judgments of its realism and we compute HYPE-Style = $\sum_i l_i^s$ for each style $s$, summing across images of that particular style. Thus, higher scores on generated images indicate higher fool rates and seem more realistic to humans on average. We use these style-level scores as the human baseline, where higher scores indicate more realistic styles, which we call \textit{HYPE-Style}. This human evaluation, while more precise and reliable, is expensive and time-consuming to perform per style: we thus set out to find automated methods that are most correlated with human judgment to assess a much larger set of styles than is cost-efficient for HYPE-Style.

\subsection{Automated Style Ranking Methods} \label{style}

We adapted FID and KID to compute distances between real and generated distributions within a single style, and use these as the style scores. We also experimented with different layers of Inception-V3 trained on ImageNet~\cite{inception} that span low-level (pool~1) to high-level (pool~3) features. For our evaluation, we included features from all three pooling layers, as well as the feature map before the auxiliary classifier (pre-aux). In total, we evaluate eight automated methods \{FID, KID\} $\times$ \{pool~1, pool~2, pre-aux, pool~3\}.

As shown in Table~\ref{result-table}, both FID and KID using pre-aux embeddings exceed the other metrics in correlating with human HYPE-Style scores, with a moderate correlation ($r$=0.433 and $r$=0.432, respectively). Following these metrics, in third rank order is: (2) FID using pool~3 embeddings ($r$=0.407), or the original FID score, then (2) KID using pool~3 embeddings ($r$=0.367). Finally, (3) FID and KID using pool~2 and pool~1 layers exhibit extremely weak correlation with $r < $ 0.2. When comparing performance between layers, KID and FID track each other, with pre-aux embeddings coming first, followed by pool~3, pool~2, and lastly pool~1. 

While the original FID paper proposed to use features from the third and last 2048-dimensional pooling layer (pool~3) of an ImageNet-pretrained Inception-v3 network~\cite{fid}, we find empirically that the 768-dimensional Inception-v3 layer just preceding the auxiliary classifier head (pre-aux) outperforms the pool~3 layer and other earlier pooling layers \{pool~1, pool~2\}. Intuitively, the pre-aux layer is the most feature-rich layer that is still regularized by the gradients from the auxiliary classifier. This regularization would encourage the layer to encode more general features that are less overfit to ImageNet, which is more useful on this task with a  different domain than ImageNet---ImageNet has, in fact, also been criticized for generalizing poorly to test sets in its own domain~\cite{recht2019imagenet}. We found that the choice of the pre-aux layer over pool~3 and others is consistent across FID and KID, with scores of 0.433 and 0.432 on the pre-aux layer against 0.407 and 0.367 on pool~3 for FID and KID respectively. As a note, the difference between the FID layers' $r$ values are not fully separable based on their 95\% bootstrapped CIs. 

\section{Discussion and Future Work} \label{discussion}

In this paper, we contribute a human evaluation metric for evaluating different styles on a generative model. We also evaluate eight different automated methods, finding that using Inception embeddings preceding the auxiliary classifier correlates more with human perception on this task than widely used methods using the last pooling layer. Our work is motivated largely by the dearth of available, reliable evaluation metrics for quantifying the progress of this task.

While none of the automated approaches evaluated comes sufficiently close to HYPE-Style for standalone use, our work still constitutes an initial foray into evaluating style-level attributes of multimodal cross-domain mapping, an area where it remains difficult to use mainstream automated evaluation metrics out of the box. As future work, we plan to both improve the realism of our generative model and explore better methods for evaluation. For instance, the performance of the pre-auxiliary classifier embeddings suggest that we are operating outside the domain of ImageNet, and from this insight, we are inclined to leverage other embedding spaces, e.g. the Mapillary or Cityscapes datasets~\cite{mapillary,cityscapes}, which could provide more suitable street-level scenery features that is similar to ours. The ultimate vision of this work is to create an ML architecture which, given an image from Google StreetView~\cite{streetview} based on a user-chosen location, is able to generate the most realistic image of climate-change induced extreme weather phenomena, given the contextual characteristics of that given image. While representing flooding realistically is the first step to achieve this goal, we later aim to represent other catastrophic events that are being bolstered by climate change (e.g. tropical cyclones or wildfires) using a similar approach.

\newpage
\bibliography{ganevalccai}
\bibliographystyle{ieeetr}

\end{document}